\documentclass{article}

\usepackage{microtype}
\usepackage{graphicx}
\usepackage{subfigure}
\usepackage{booktabs} % for professional tables
\usepackage{tikz}
\usetikzlibrary{fit,positioning}

\usepackage{amsmath}
\usepackage{amsfonts}
\usepackage{bm}
\usepackage{float}
\usepackage{multicol}
\usepackage{caption}

\usepackage{hyperref}

\newtheorem{theorem}{Theorem}
\usepackage[accepted]{icml2020}

\icmltitlerunning{Conjoined Dirichlet Process}

\begin{document}

\twocolumn[
\icmltitle{Conjoined Dirichlet Process}
%A Non-Parametric Probabilistic Biclustering Method


\icmlsetsymbol{equal}{*}

\begin{icmlauthorlist}
\icmlauthor{Michelle N.~Ngo}{equal,mcsb}
\icmlauthor{Dustin S.~Pluta}{equal,stat}
\icmlauthor{Alexander N.~Ngo}{sb}
\icmlauthor{Babak Shahbaba}{mcsb,stat,cs}
\end{icmlauthorlist}

\icmlaffiliation{mcsb}{Center for Complex Biological Systems, University of California at Irvine }
\icmlaffiliation{stat}{Department of Statistics, University of California at Irvine }
\icmlaffiliation{cs}{Department of Computer Science, 
 University of California at Irvine }
\icmlaffiliation{sb}{Department of Computer Science, University of California at Santa Barbara }

\icmlcorrespondingauthor{Babak Shahbaba}{babaks@uci.edu}

\icmlkeywords{Machine Learning, Clustering, Infinite Mixture Model, Dirichlet Process, ICML}

\vskip 0.3in
]



\printAffiliationsAndNotice{\icmlEqualContribution} % otherwise use the standard text.

\begin{abstract}
Biclustering is a class of techniques that simultaneously clusters the rows and columns of a matrix to sort heterogeneous data into homogeneous blocks. 
Although many algorithms have been proposed to find biclusters, existing methods suffer from the pre-specification of the number of biclusters or place constraints on the model structure. 
To address these issues, we develop a novel, non-parametric probabilistic biclustering method based on Dirichlet processes to identify biclusters with strong co-occurrence in both rows and columns. 
The proposed method utilizes dual Dirichlet process mixture models to learn row and column clusters, with the number of resulting clusters determined by the data rather than pre-specified. 
Probabilistic biclusters are identified by modeling the mutual dependence between the row and column clusters. 
We apply our method to two different applications, text mining and gene expression analysis, and demonstrate that our method improves bicluster extraction in many settings compared to existing approaches.
\end{abstract}

\section{Introduction}
\label{introduction}

%\medskip

Biclustering, or co-clustering, is a technique used for sorting heterogeneous data into homogeneous blocks by allowing for simultaneous clustering of the rows and columns of a matrix. 
This technique has various important applications, including text mining and biological gene expression analysis. 
In text mining, biclustering text data from a document corpus allows for identification of document-word combinations with high co-occurrence. 
Extracted biclusters represent combinations of words and documents that form a (latent) topic.
Biclustering has been particularly popular in the past several decades for gene expression microarray analyses. 
The method is used to group genes into similar conditions to study the functional roles of genes. 
More recently, biclustering is being used to analyze single cell RNA sequencing data. 
Here, the method is usually used to study cell proliferation by grouping cells into developmental stages and identifying the genetic drivers for each stage.

Current biclustering methods generally impose restrictive assumptions on the biclustering structure or data-generating mechanisms.  
However, in real-world applications, which are often exploratory, an appropriate model and bicluster structure can be difficult to specify.  
To address these limitations in current methods, we propose the Conjoined Dirichlet Process (CDP): a novel, non-parametric probabilistic biclustering method based on dual Dirichlet processes to identify biclusters with strong co-occurrences in both rows and columns. 
The name of the method derives from its usage of two conjoined DPMMs, akin to conjoined twins (see Figure \ref{fig:plate}). 
CDP provides the following advantages:
1) the number of biclusters is determined by the data and prior, and does not require selecting a number of clusters \textit{\'a priori}, 2) fewer modeling assumptions compared to parametric alternatives, 3) estimated biclusters may overlap arbitrarily, and 4) efficient computational methods allow applications to high dimensional data, making applications to text and gene expression data practical. 

The paper is organized as follows. 
In Section \ref{sec:previousmethods} we describe existing biclustering methods. 
In Section \ref{sec:background} we provide some background on Dirichlet process mixture models (DPMMs), particularly focusing on the parallel MCMC sampler for DPMMs. 
In Section \ref{sec:CDP} we discuss and provide details of our proposed biclustering method. 
In Section \ref{sec:results} we apply our method to simulated, text, and single cell RNA sequencing data sets, and present the results. 
Finally, in Section \ref{sec:discussion} we present our conclusion.
%%%
%In text mining, biclustering text data from a document corpus allows for identification of document-word combinations with high co-occurrence. 
%Extracted biclusters represent combinations of words and documents that form a (latent) topic.  
%This is distinguished from traditional latent Dirichlet allocation (LDA) \cite{Blei2003LatentAllocation} topic modeling in that LDA does not cluster documents directly, and words which co-occur across many documents may be clustered even if the share vocabulary of those documents is small overall.
%Instead, biclustering may encourage heavy topics which exhibit high co-occurrence of words across documents and documents across words.

%In biological gene expression analyses, biclustering has been a popular technique for analyzing gene microarray data. 
%Particularly, biclustering allows for the grouping of similar genes under multiple conditions and conditions under multiple genes, without assigning equal weight to each gene or each condition. 
%The ability of overlapping biclusters, indicating that a gene may be involved in multiple conditions or vice versa, is also biologically relevant. 
%More recently, biclustering is also  being applied to single cell RNA sequencing (scRNA seq) data. 
%Biclustering scRNA seq data is used to define developmental stages based solely on the transcriptome in addition to accounting for variation in the data, and identifying biologically important genes and their signatures for each cell stage. 
%Each bicluster is an association between groups of cell stages and their genetic drivers. 
%%%

\section{Previous Methods} \label{sec:previousmethods}
 
Briefly, biclustering algorithms are based on four heuristics: greedy, divide-and-conquer, exhaustive enumeration, or distribution parameter identification \cite{Padilha2017ATechniques}. 

\cite{Hartigan1972} proposed the first biclustering algorithm in 1972, but the technique was not popular until 2000 when \cite{Cheng2000BiclusteringData} applied it to gene microarray data.  
%Theirs is a greedy approach; the method starts with the largest possible biclusters and removes rows and columns to find biclusters with the smallest mean squared residue (a measure of bicluster homogeneity). 
%The mean squared residue measures how each element in a bicluster differs from the row mean, column mean, and mean of the bicluster. 
Other popular gene microarray biclustering algorithms include \cite{Kluger2003SpectralConditions}'s spectral model and \cite{Lazzeroni2002}'s plaid model.
%Both models use a distribution parameter identification approach, where a proposed statistical model captures the structure of the biclusters, and the algorithms iteratively updates the model parameters. 
%The spectral model uses singular value decomposition to simultaneously cluster rows and columns; the resulting data matrix has a checkerboard pattern. 
%The plaid model uses binary least squares to fit the membership parameters. 
%Each parameter in this model describes one "layer" of the plaid image such that the gene expression value in a bicluster is the sum of the "average" effect, gene effect, cell effect, and some noise. 

While many biclustering algorithms have been developed for gene microarray analyses, one of the first applications for biclustering was text mining.
Dhillon et al proposed two different biclustering algorithms for simultaneously partitioning documents and words: spectral co-clustering \cite{dhillon2001co}, and a co-clustering algorithm based on information theory \cite{Dhillon2003}. 
%Spectral co-clustering is based on partitioning bipartite spectral graphs using singular vectors of a matrix; 
\cite{Kluger2003SpectralConditions}'s spectral model for gene microarray analyses is based on \cite{dhillon2001co}'s spectral model. 
%Dhillon's second algorithm attempts to find biclusters that maximize the mutual information between clusters. 

More recently, biclustering has been applied to single cell RNA sequencing (scRNA seq) data. 
Biclustering methods specific to this application include BackSPIN \cite{Zeisel2015} and QUBIC2 \cite{Xie2019QUBIC2}. 
%BackSPIN \cite{Zeisel2015}, which the authors used to discover distinct cell classes, is based on sorting observations into neighborhoods i.e. by gene-gene or cell-cell similarity. 
%QUBIC2 is a statistical biclustering model, utilizing a truncated Gaussian mixture model and Kullback-Leiber divergence score optimization to identify condition-specific functional gene modules \cite{Xie2019QUBIC2}. 

\cite{Rugeles2017Biclustering:Models} developed Dual Topics for Bicluster (DT2B), a biclustering method based on a generalized latent Dirichlet allocation (LDA) model \cite{Blei2003LatentAllocation}. Unlike the previous models, DT2B avoids the constraints of a model structure.
However, the algorithm requires a discretized data set, pre-specification of the number of row and column clusters, and threshold values. 
 
By using a Dirichlet process mixture model (DPMM) instead of a LDA model, we bypass the need to specify the number of biclusters, make strong modeling assumptions, and particular data format. 

\section{Background} \label{sec:background}

\subsection{Latent Dirichlet Allocation}
Latent Dirichlet allocation (LDA) is a hierarchical Bayesian model used to infer latent features in collections of discrete data. 
Initially proposed to estimate and describe population structure from genotype data \cite{pritchard2000inference}, it is also commonly used for the classification of documents based on word frequencies \cite{Blei2003LatentAllocation}.  

In the context of document classification, LDA posits that for a corpus of documents, the probability distribution of words for a given document is determined by a set of latent "topics" associated with that document.  
LDA infers these latent topics from observed word frequencies for each document to produce a clustering or classification of documents in the corpus. 
\vspace{-0.2cm}

\begin{figure*}[ht!]
\centering
\includegraphics[scale=0.883]{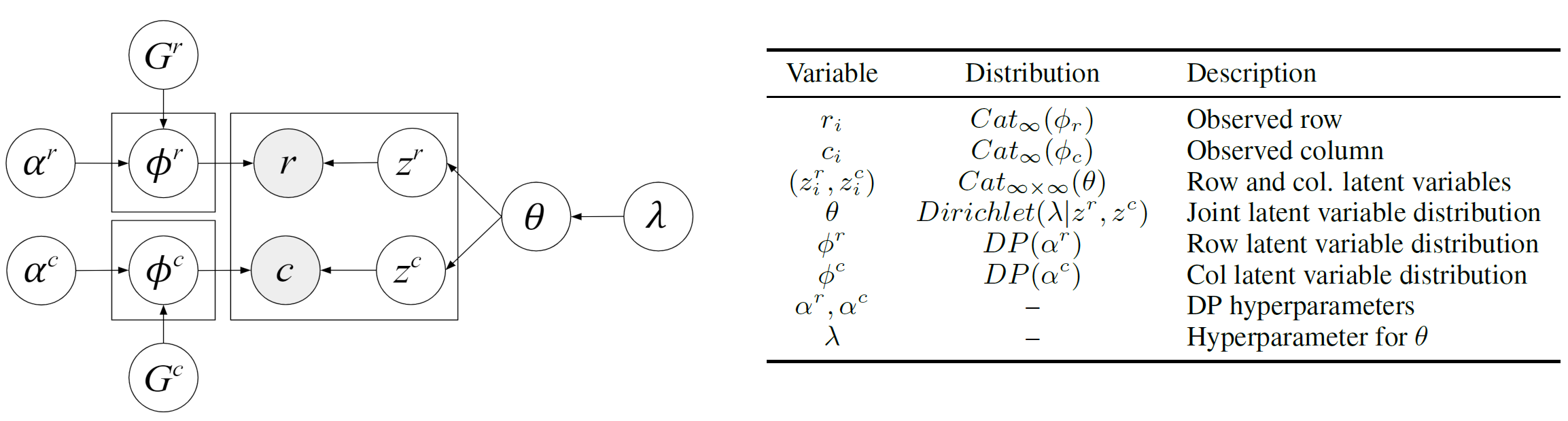}
\caption{Plate diagram and description of included variables and parameters.  Rows $r$ and columns $c$ are clustered separately through the DPMMs defined by $\phi^r$ and $\phi^c$.  After updating according to Algorithm 1, heavy biclusters can be extracted from $\phi^r, \phi^c$, and $\theta$ (the joint distribution of latent cluster assignments given by $z^r, z^c$).}
\label{fig:plate}
\end{figure*}

\subsection{Dirichlet Process Mixture Model}
DPMMs remove the need to pre-specify the number of clusters by placing a Dirichlet process (DP) prior over the cluster parameters, and in this sense, allows ``infinite'' mixture models to incorporate automatic model selection. 

The DPMM is intuitively an infinite dimensional generalization of a mixture of Dirichlet distributions. 
We begin by considering a Bayesian mixture model with $K$ clusters and then extending $K \rightarrow \infty$: 
\begin{align*}
x_i | z_i, \theta_i &\sim F(\theta_{c_i}) \\
z_i | \bm{\pi} &\sim Discrete(p_1, \dots, p_K) \\
\theta_c &\sim G_0 \\
\bm{\pi} &\sim Dirichlet \left(\frac{\alpha}{K}, \dots, \frac{\alpha}{K} \right) 
\end{align*}
Here $x_1, \dots, x_n$ is the observed data and drawn from a mixture of distributions with the form $F(\theta)$, $\theta$ is the mixing distribution over $G$, and $z$ is the cluster assignments for each observation \cite{Neal2000MarkovModels}.
The prior for our mixing distribution is a Dirichlet process with concentration parameter $\alpha$ and base distribution $G_0$. 
For a more in-depth explanation, see \cite{Sudderth2006GraphicalTracking}. 

\subsection{Parallel Sampling of DPMMs}
DPMMs have been largely computationally heavy to implement.
\cite{Chang2013ParallelSplits} parallelized the MCMC sampler for DPMMs by utilizing a restricted Gibbs sampler to fix the number of clusters before proposing splits and merges. 
Since the number of clusters are fixed, each of the Gibbs sampler steps can be done in parallel.
Furthermore, to increase efficient cluster splits, they augment each cluster with two sub-clusters, labeled $\bar{z}_i \in \{l, r\}$ to denote whether each data point $x_i$ is associated with the left or right sub-cluster.
Additional auxiliary variables introduced are the sub-cluster weights $\bar{\pi}_{k} \in \{\bar{\pi}_{k,l}, \bar{\pi}_{k,r}\}$ and parameters $\bar{\theta}_{k} \in \{\bar{\theta}_{k,l}, \bar{\theta}_{k,r}\}$ of cluster $k$. 
The auxiliary variables for the sub-clusters are analogous in function to the variables for the regular clusters. 
In this augmented restricted Gibbs sampling algorithm, we now sample a regular cluster assignment and then a sub-cluster assignment for each data point.
Splits and merges, to either split a cluster into its two sub-clusters or merge two sub-clusters into one new cluster, are proposed and accepted with probability $\min(1, H)$, where $H \in \{H_{split},H_{merge}\}$ is the Hastings ratio for the respective action.

\cite{Dinari2019DistributedJulia} extended this implementation to enable parallelization on multiple multi-core machines instead of a single multi-core machine.
The authors note that sampling cluster parameters $\theta_k$ is parallelizable over the clusters, sampling cluster assignments $z_i$ is independently computed for each data point $x_i$, and proposing cluster splits is parallelizable. 
For computational efficiency, they rely on a distributed-memory model and utilize sufficient statistics to communicate between the cores as well as the between the machines. 
The sufficient statistic $T$ for a multinomial cluster (e.g. for document classification or single cell RNA sequencing data analysis) is $T = \sum_{i=1}^{N} x_i \in \mathbb{N}_0^d,$ where $d$ is the dimension of the data points $x_i$. 
The aggregation of the sufficient statistics for each cluster allows for the sampling of cluster parameters across multiple parallelized worker processes.
Splits and merges are proposed similarly to \cite{Chang2013ParallelSplits} on the master process, with mappings of old cluster assignments to new assignments broadcasted to all worker processes to individually update its data points. 
Using this multi-machine, multi-core implementation considerably speeds up our model and allows us to handle high dimensional data. 

\section{Conjoined Dirichlet Process (CDP)} \label{sec:CDP}

CDP is a probabilistic biclustering method that provides several important characteristics in the context of gene-cell count analysis.  
The estimated biclusters may overlap, posterior probabilities of each element belonging to a given bicluster can be calculated, and heavy biclusters (showing strong co-occurrence in rows and columns) are encouraged.  
By utilizing a pair of DPMMs for bicluster estimation, CDP eliminates the need to specify the number of row topics and columns topics \textit{\'a priori}, which is particularly relevant for both gene expression and document analysis where the number of topics and biclusters is unknown or ill-defined.

\subsection{Model Construction}
CDP can be summarized in two steps:
\vspace{-0.3cm}
\begin{enumerate}
    \item Use DPMMs to learn row and column clusters. \vspace{-0.2cm}
    \item Model the mutual dependence between the row and column clusters to extract biclusters with strong co-occurrence values in both rows and columns.
\end{enumerate}
%The construction is based on DT2B \cite{Rugeles2017Biclustering:Models}; however DPMMs are used to update the row-cluster assignments $z^r$ and column-cluster assignments $z^c$ instead of LDA. 
\vspace{-0.2cm}

\begin{figure*}[ht!]
\centering
\includegraphics[scale=0.125]{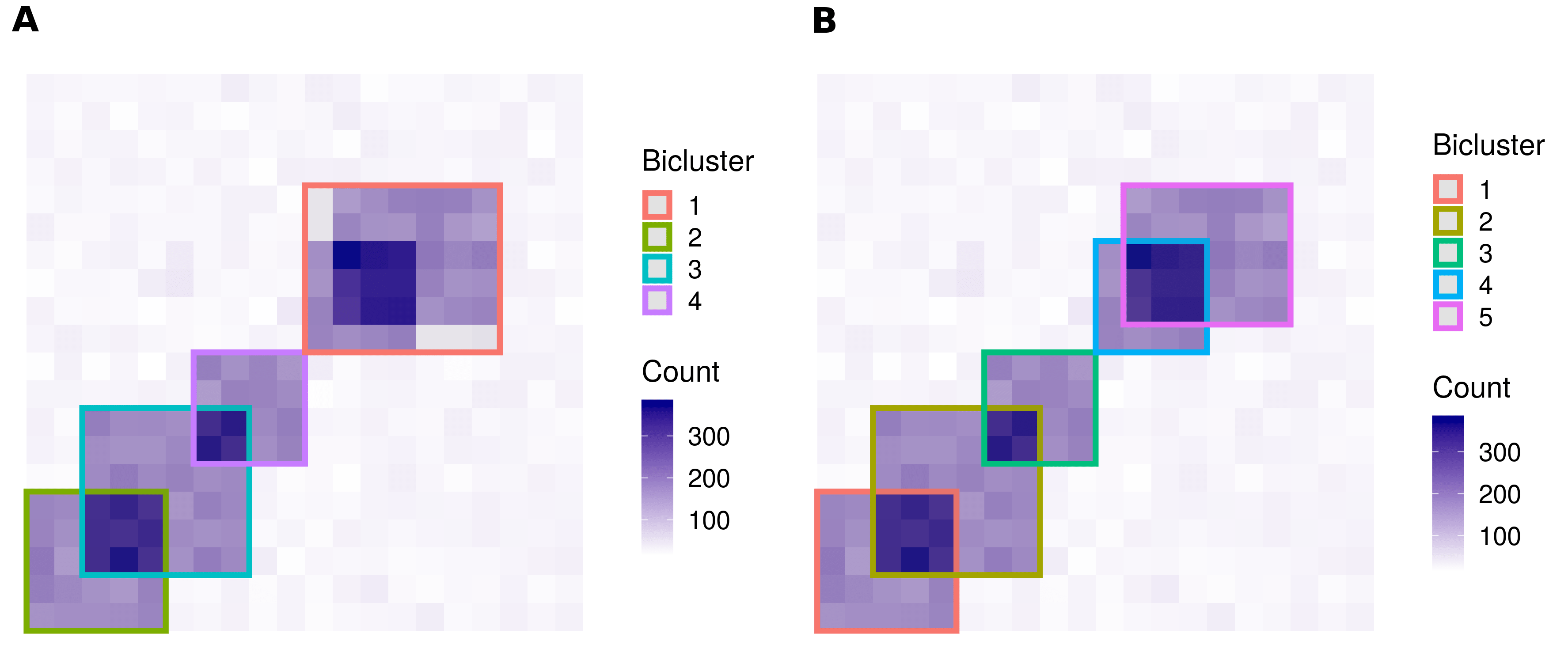}
\caption{CDP is able to detect overlapping biclusters: (A) Heatmap of simulated count data and bicluster membership estimated by CDP. (B) True bicluster structure for simulated data.}
\label{fig:illustrative}
\end{figure*}

Given a $n_R \times n_C$ matrix where $n_R$ is the number of rows and $n_C$ is the number of columns, each matrix entry $(r, c)$ represents the frequency of row $r$ in column $c$. 
For text data, this corresponds to the frequency of word $r$ in document $c$ and for single cell RNA sequencing data, this corresponds to the gene expression of gene $r$ in cell $c$. 

Using a DPMM, we can sequentially cluster the rows and columns of the matrix to obtain row-cluster assignments $z^r$ and column-cluster assignments $z^c$. 
Similar to DT2B \cite{Rugeles2017Biclustering:Models}, we now have two sets of latent variables (e.g. topics for text data) and use these sets to extract biclusters with strong co-occurrence values in rows and columns. 

Figure~\ref{fig:plate} shows the graphical model for CDP, where row $r$ and column $c$ are the rows and columns of the data matrix. 
$z^r$ and $z^c$ are the vectors of row and column cluster indices (assignments) respectively. 
$\phi^r$ is the row per row latent variable distribution, $\phi^c$ is the column per column latent variable distribution, and $\theta$ is the joint latent variable distribution. 
These three variables maintain the counting over the relationships between the data, latent variables and their mutual dependence.
For discrete data, the hyperparameters for CDP are $\gamma$, the concentration parameter for the DP; $\beta$, the prior for the DP measure; $\alpha^r$, the hyperparameter for $\phi^r$; $\alpha^c$, the hyperparameter for $\phi^c$; and $\lambda$, the hyperparameter for $\theta$. Figure~\ref{fig:illustrative} shows an illustrative example of CDP. 
\begin{theorem}
If row assignments $z^r$ are held fixed, then the CDP update step is equivalent to a latent Dirichlet allocation update on $z^c$.  A similar result holds if $z^c$ is held fixed for updating $z^r$. 
\end{theorem}
\noindent \textit{Proof:} In evaluating Eq. \ref{eqn:phic} to update $z^c$, we can then treat $\phi^r$ as a constant, yielding
\begin{align}
    P(z_i^c = j &| c_i = m, r_i = n, \mathbf{z}_{-i}^c) \\
    &\propto \phi^c_{mj}\theta_{jk}\\
    &\propto \frac{C_{mj} + \alpha^c}{\sum_{m'} C_{m'i} + n_C\alpha^c}(C_{ij} + \lambda).
\end{align}
Updating $z^c$ according to this probability is equivalent to the update given by LDA \cite{Blei2003LatentAllocation}.

\subsection{Inference Process}
Algorithm~\ref{alg:CDP} shows the inference process for CDP, using the distributed MCMC inference algorithm outlined in \cite{Dinari2019DistributedJulia}, which is based on the restricted Gibbs sampler method in \cite{Chang2013ParallelSplits}. 
Due to the split and merge aspect and the high-dimensionality of our data, the posterior distribution of the assignment parameters may be multi-modal. 
For this reason, we update the assignment parameters for a specified number of iterations and take the $MAP$ estimate of the maximum values of $z^r$ and $z^c$.  

The specifications for the hyperparameters of CDP (under the assumption that the base distribution of the DP is multinomial) are listed below: 
\begin{multicols}{2}
\noindent
\begin{align*}
\gamma^r, \gamma^c &\in \mathbb{R}^{1 \times 1} \\
\beta^r &\in \mathbb{R}^{n_R \times 1} \\
\beta^c &\in \mathbb{R}^{n_C \times 1} 
\end{align*}
\begin{align*}
\alpha^r &\in \mathbb{R}^{K_r \times 1} \\
\alpha^c &\in \mathbb{R}^{K_c \times 1} \\
\lambda &\in \mathbb{R}^{K_r \times K_c} 
\end{align*}
\end{multicols}
\vspace{-0.5cm}
Given a collection of composites (e.g. documents, cells) $C$ made up of parts (e.g. words, genes) $R$, we can write the probability of a composite $c$ containing a part $r$ as: 
\begin{align*}
P(r, c) &= \sum_{z_r} \sum_{z_c} P(r | \phi_{z_r}^r, \alpha^r) P(c | \phi_{z_c}^c, \alpha^c) P(z_r, z_c | \theta)
\end{align*}
A major advantage of the CDP over DT2B \cite{Rugeles2017Biclustering:Models} is that the CDP does not require thresholds to control the trade-off between quantity and quality of the biclusters. 
The hyperparameters in the DPMM step facilitate this trade-off automatically. 
Setting a large $\gamma$, the Dirichlet process concentration parameter, and for a multinomial base distribution, a large $\beta$ (the Dirichlet distribution hyperparameter) will yield more  clusters. 

The probabilistic biclusters are given by the joint distribution of row and column latent variables, $\theta$, which has dimension $K_r \times K_c$. 
$K_r$ and $K_c$ are the number of latent row and column variables respectively.
As previously mentioned, from the DPMMs, we obtain the row-cluster assignments $z^r$ and column-cluster assignments $z^c$.  
Calculating the mode of the posterior distributions of $z^r$ and $z^c$ yields the maximum a posteriori (MAP) estimate of the number of latent row and column variables, i.e. $K_r$ and $K_c$. 

We note that the dimensions of the row per row latent variable distribution, $\phi_r$, and the column per column latent variable distribution, $\phi_c$, are also given by the MAP. 
$\phi_r$ has dimensions $n_R \times K_r$, and $\phi_c$ has dimensions $n_C \times K_c$. 

\setlength{\textfloatsep}{4pt}
\begin{algorithm}[htb]
	\caption{Conjoined Dirichlet Process (CDP)}
	\label{alg:CDP}
	\begin{algorithmic}
		\STATE {\bfseries Input:} Data $\bm{X}$, size $n_R \times n_C$
		\STATE \hskip3em DP concentration parameters $\gamma_R, \gamma_C$
		\STATE \hskip3em Dirichlet distribution hyperpriors $\beta_R, \beta_C$
		\STATE \hskip3em Number of DPMM iterations $iter_R$, $iter_C$
		\STATE \hskip3em Number of cluster reassignment iterations $iter_U$
		\FOR{$i=1$ {\bfseries to} $iter_R$}
		\STATE Run DPMM on $\bm{X}$
		\ENDFOR
		\FOR{$i=1$ {\bfseries to} $iter_C$}
		\STATE Run DPMM on $\bm{X}^T$
		\ENDFOR
		\STATE Calculate $K_r = MAP(z^r)$ and $K_c = MAP(z^c)$
		\FOR{$i=1$ {\bfseries to}  $iter_U$}
		\STATE Update $z^r$ and $z^c$ using the data as weights
		\ENDFOR
		\FOR{$i=1$ {\bfseries to} $n_R$}
		\FOR{$j=1$ {\bfseries to} $K_r$}
		\STATE Calculate $\phi_{ij}^r = \frac{C_{ij} + \alpha^r}{\sum_{i'} C_{i'j} + n_R \alpha^r}$
		\ENDFOR
		\ENDFOR
		\FOR{$i=1$ {\bfseries to} $n_C$}
		\FOR{$j=1$ {\bfseries to} $K_c$}
		\STATE Calculate $\phi_{ij}^c = \frac{C_{ij} + \alpha^c}{\sum_{i'} C_{i'j} + n_C \alpha^c}$
		\ENDFOR
		\ENDFOR
		\STATE Calculate $\theta \propto C_{r,c} + \lambda$, $1 \leq r \leq n_R, 1 \leq c \leq n_C$ 
	\end{algorithmic}
\end{algorithm}

\subsection{Bicluster Extraction}
From the DPMM, we obtain latent variables $z^r$ and $z^c$, which indicate the row and column cluster assignments respectively. 
To extract the biclusters from the data, we need to calculate three parameters: row per row latent variable distribution $\phi^r$, column per column latent variable distribution $\phi^c$, and joint distribution of row and column latent variables $\theta$. 

These three parameters are given by \cite{Rugeles2017Biclustering:Models}: 
\begin{align}
\phi_{mi}^c &= \frac{C_{mi} + \alpha^c}{\sum_{m'} C_{m'i} + n_C \alpha^c} \label{eqn:phic} \\
\phi_{nj}^r &= \frac{C_{nj} + \alpha^r}{\sum_{n'} C_{n'j} + n_R \alpha^r} \label{eqn:phir}\\
\theta &\propto C_{ij} + \lambda \label{eqn:theta}
\end{align}
where $C_{ab}$ is the number of instances $a$-th variable is assigned to $b$-th variable. 
For example, $\phi^r$ is the probability of the $n$-th row being assigned to $j$-th row latent variable. 
Thus, $C_{nj}$ is the number of times the $n$-th row is assigned to to $j$-th row latent variable. 
The joint distribution $\theta$ tracks the relationship between the current row and column latent variables to capture the mutual dependence between the two sets of latent variables \cite{Rugeles2017Biclustering:Models}. 

First, we calculate $\phi^r$ and $\phi^c$ by using the aforementioned sets of latent variables $z^r$ and $z^c$ as the initial cluster assignments. 
These assignments are updated iteratively using the data as weights. 
Once the row and column assignments have been updated, we count the number of instances a row or column is assigned to a row or column latent variable. 

To obtain the joint distribution of row and column latent variables $\theta$, we need to calculate the frequency of each row and column latent variable pairing $(i, j)$. 
The vector of frequencies for each row and column latent variable pairing is then transformed into a contingency table of size $K_r \times K_c$, i.e. the desired $\theta$. 

\subsection{Implementation Overview}
To obtain the row-cluster assignment $z^r$ and column-cluster assignment $z^c$, we separately infer each parameter using \cite{Dinari2019DistributedJulia}'s implementation in Julia. 
We utilize a specific version of that package that outputs the cluster assignments $z^r$ or $z^c$ at each iteration rather than the final cluster assignments. 
While this requires more memory storage and run time, it allows CDP to have overlapping biclusters and more interpretable results depending on the application. 

For $N$ data observations, $K$ clusters, and $M$ machines with $P$ cores, the total runtime complexity for the DPMM implementation is $\mathcal{O}(K) + \mathcal{O}(M + P) + \mathcal{O}(N K / (M P) )$. 
For more details on the runtime complexity for the DPMM, see \cite{Dinari2019DistributedJulia}. 

CDP reassigns each observation iteratively in batches of size equal to either the row sums or column sums. 
These batches are parallelized to run on $P$ processes (cores). 
Thus, updating the row and cluster assignments for $J$ iterations takes $\mathcal{O}(N J / P)$ time.
Calculating $\phi$ for the rows and columns require the aforementioned assignment step. 
Once reassigned, CDP splits the $N$ data points into vectors of length row sums (for $\phi^r$) or column sums (for $\phi^c$). 
These vectors are then tabulated over the number of latent variables $K$ to determine the probability of each row or column being assigned to each row latent variable or column latent variable respectively. 
The runtime complexity for calculating $\phi$ excluding the cluster assignment update step is then $\mathcal{O}(N K)$ where $K$ is equal to $K_r$ when calculating $\phi^r$ and $K_c$ when calculating $\phi^c$. Calculating the joint distribution of both row and column latent variables $\theta$ requires looping over the assignments for one direction (e.g. row assignments) and matching the row and column indexes to the assignments in the other direction (e.g. column assignments). 
This operation requires $\mathcal{O}(N)$ time. 
CDP then tabulates the row and column assignment of each row and column pairing to obtain $\theta$. 
Thus, the total runtime complexity for calculating $\theta$ is $\mathcal{O}(N K_r K_c)$ and $\mathcal{O}(N K_r K_c / P)$ if run in parallel. 
  
As $N \gg K, P, M$ and $J$, CDP takes $\mathcal{O}(NJ) + \mathcal{O}(N K_r K_c)$ time. 
Experiments were conducted on an i5-7600K CPU. 

\section{Experimental Results} \label{sec:results}

We compare CDP to DT2B \cite{Rugeles2017Biclustering:Models} because this method also models the mutual dependency between two sets of latent variables. 
We also compare our algorithm to spectral biclustering \cite{Kluger2003SpectralConditions} since both try to extract high co-occurences. 
For completeness, Cheng and Church \cite{Cheng2000BiclusteringData} and the plaid \cite{Lazzeroni2002} algorithms are also used for comparisons due to their common usage, and BiMax \cite{Prelic2006BiMax} which is known to serve as a reference method.

\subsection{Data sets}

\subsubsection{Synthetic Data}

\begin{figure}[h!]
    \centering
    \includegraphics[width=0.95\columnwidth]{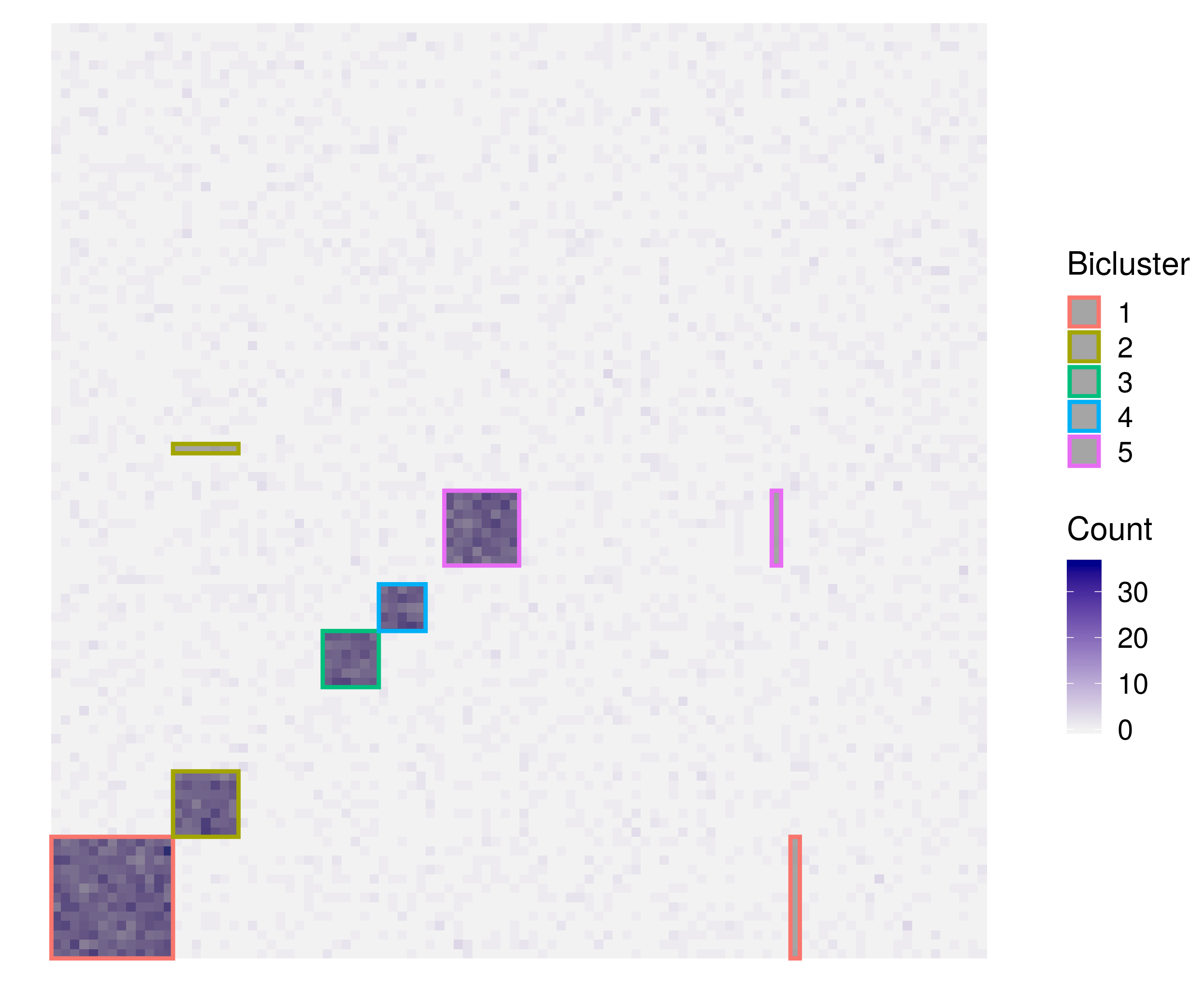}
    \caption{Example results from CDP for simulated data (case 4).  CDP correctly identifies the heavy biclusters (0.938 Jaccard score), with only a small number of spurious elements included (e.g. in biclusters 1, 2 and 5). The data shown here is approximately 70\% sparse.}
    \label{fig:simulated_biclusters}
\end{figure}

Simulated count data were generated from a multinomial distribution defined by an $R \times C$ probability matrix $\theta$ (with entries summing to 1), and by fixing the sum of entries in the resulting random matrix at some total count $N$.  
The total bicluster probability $p$ of an element belonging to a bicluster was set to control the strength of biclusters and overall sparsity.  
Four different constructions of $\theta$ were chosen to evaluate performance over different biclustering patterns.  
In order of increasing complexity, these four cases are (1) a single distinct bicluster, $N = 4000, R = C = 50, p = 0.8$; (2) two distinct biclusters, $N = 4000, R = C = 20, p = 0.5$; (3) 3 biclusters with one overlap $N = 4000, R = C = 50, p = 0.7$; (4) 5 distinct biclusters, $N = 10000, R = C = 100, p = 0.7$ (see Figure \ref{fig:simulated_biclusters} for an example). 

To compare the performance of CDP to existing methods, we use the Jaccard score, defined as
$J(\mathcal{B}_1, \mathcal{B}_2) = \min_{(\mathcal{A}, \mathcal{B})} \frac{1}{|\mathcal{A}|}\sum_{A \in \mathcal{A}} \max_{B \in \mathcal{B}} \frac{|A \cap B|}{|A \cup B|},$ where $\mathcal{B}_1, \mathcal{B}_2$ are two sets of biclusters, with the minimum taken over $(\mathcal{A}, \mathcal{B}) \in \{(\mathcal{B}_1, \mathcal{B}_2), (\mathcal{B}_2, \mathcal{B}_1)\}$.  
The Jaccard score is a symmetric similarity metric taking values $0 \leq J(\mathcal{B}_1, \mathcal{B}_2) \leq 1$, with the lower bound attained only when all sets in $\mathcal{B}_1$ are disjoint with all sets in $\mathcal{B}_2$ and the upper bound attained only when $\mathcal{B}_1 = \mathcal{B}_2$.  
In the context of the simulation study, $\mathcal{B}_1$ is the set of estimated biclusters from a given method, and $\mathcal{B}_2$ is the set of true biclusters from the generative model.

\subsubsection{Real-life Data}
\begin{enumerate}
    \item \textbf{Condensed 20 Newsgroups}: Collection of 100 words across 16,242 newsgroup documents ("netnews"). The data is organized into 17 different newsgroups and 4 main topics. This data set is $95.97\%$ sparse. 
    %\vskip 0.4in
    \item \textbf{Single cell RNA sequencing (scRNA seq) Data}: Collection of 23,226 genes across 5,053 transcriptomes from 10 distinct regions of murine juvenile and adult central nervous system \cite{Marques2016OligodendrocyteSystem.}. All cells were profiled using the Fluidigm C1 system and sequenced on an Illumina HiSeq 2000 instrument. This data set is $87.57\%$ sparse.
\end{enumerate}

\subsection{Parameter Settings}
For CDP, we need to set the number of iterations, the Dirichlet process concentration parameter $\gamma$, and the Dirichlet distribution hyperprior $\beta$. 
Note that both our text and biological data are discrete counts so we assume a multinomial base distribution.
If we had continuous data we would instead assume a Gaussian base distribution (or another continuous distribution) and set the values for a Normal--Inverse--Wishart hyperprior. 
The hyperparameters for $\phi^r$, $\phi^c$ and $\theta$ are set to zero by default. 
We set all concentration parameters and hyperpriors to be small to obtain larger cluster sizes. 
Table~\ref{tab:parameters} shows the parameter values for the two real data sets. 
We did not include the two $\beta$ hyperparameters or the $\lambda$ hyperparameter in the table since we set those values to zero. 
In practice, if one has strong prior knowledge regarding a row or column element, setting a value greater than zero for those hyperparameters will result in a more accurate clustering.
However, we are doing strictly exploratory work for this paper.
\vskip -0.2in
\begin{table}[H]
	\caption{Parameter settings for the DPMM part of CDP on two data sets.}
	\label{tab:parameters}
	\vskip 0.15in
	\begin{center}
		\begin{small}
			\begin{sc}
				\begin{tabular}{lccccr}
					\toprule
					Data set & Row/Col & Iterations & $\gamma$ & $\beta$ \\
					\midrule
					Newsgroups     & Row & 1000  & 10    & 1 \\
					               & Col & 1000 & 100    & 1 \\
					scRNA Seq      & Row & 500 & 10 & 0.1 \\
					               & Col & 500  & 10  & 1    \\
					\bottomrule
				\end{tabular}
			\end{sc}
		\end{small}
	\end{center}
	\vskip -0.1in
\end{table}

\begin{table*}[ht!]
\caption{Comparison of mean (standard deviation) Jaccard similarity scores for various biclustering algorithms on the simulated data sets.}
\label{tab:comparisons_simulated}
\vskip 0.15in
\begin{center}
\begin{small}
\begin{sc}
\begin{tabular}{lccccc}
\toprule
Case & Plaid & C \& C & Bimax & DT2B & CDP\\
\midrule
1 & 0.133 (0.06) & 0.16 (0.00) & 0.141 (0.005) & 0.806 (0.204) & 0.69 (0.088)\\
2 &  0.862 (0.265) & 0.016 (0.000) & 0.098 (0.021) & 0.889 (0.118) & 0.985 (0.081)\\
3 & 0.172 (0.087) & 0.048 (0.000) & 0.105 (0.012) & 0.236 (0.032) & 0.25 (0.014)\\
4 & 0.316 (0.343) & 0.008 (0.000) & 0.101 (0.014) & 0.522 (0.087) & 0.756 (0.033) \\
\bottomrule
\end{tabular}
\end{sc}
\end{small}
\end{center}
\vskip -0.1in
\end{table*}

\subsection{Results for Synthetic Data}

Results for the four synthetic data cases are provided in Table \ref{tab:comparisons_simulated}. In each of the cases considered, plaid, DT2B, and CDP exhibit the highest accuracy in bicluster estimation as measured by the Jaccard score.  We also tested the spectral method, but the accuracy was so low we excluded it from the table. In cases 2, 3, and 4, CDP outperforms all other methods, and gives substantially better performance in the most complicated setting (case 4), with a mean Jaccard similarity of 0.756, compared to DT2B with a mean score of 0.522.  CDP also exhibits lower variance over repeated simulations compared to DT2B.  Only in the simplest setting of a single bicluster (case 1) does DT2B show better mean similarity score, with 0.806 for DT2B compared to 0.69 for CDP.  However, DT2B shows high variance in the accuracy of its estimates over repeated runs in this case, whereas CDP shows lower variance over all scenarios.  Together, these results suggest that CDP is practical for bicluster extraction, and may be significantly more accurate compared to existing methods.

\subsection{Simulation Runtime Comparisons}
The DT2B method is conceptually similar to CDP, but requires selecting a maximum number of row and column clusters to determine the parameters of the underlying LDA models.  In general, DT2B is most efficient and accurate when the maximum number of row clusters ($K_r$) and column clusters ($K_c$) are set to the true number of row and column clusters respectively, but these values will be unknown in practice.  DT2B runs in $O(N K_r K_c)$ time \cite{Rugeles2017Biclustering:Models}, thus setting $K_r$ and $K_c$ to the number of rows and columns respectively may be computationally prohibitive for applications to single cell analysis and other large data settings.  Table 
\ref{tab:runtime_comp} shows the runtime of DT2B for different choices of $K_r$ and $K_c$ on a simulated data set, compared to CDP.  

\begin{table}[htb]
    \centering
    \caption{Comparison of runtimes for CDP and DT2B with various choices of $(K_r, K_c)$ on a simulated data set (case 2).  Runtimes for DT2B scale linearly in both $K_r$ and $K_c$.}
    \begin{tabular}{lcc}
    \toprule
    Method & Mean Jaccard (s.d.) & Runtime (s)\\
    \midrule
    CDP & 0.96 (0.02) & 13.22\\
    DT2B(5, 5) & 0.41 (0.11) & 4.23\\ 
    DT2B(10, 10) & 0.94 (0.13) & 12.85\\ 
    DT2B(25, 25) & 0.98 (0.01) & 73.45\\ 
    \bottomrule
    \end{tabular}
    \label{tab:runtime_comp}
\end{table}

\subsection{Results for Text Data}
Biclustering text data from a document corpus allows for identification of document-word combinations with high co-occurrence.  Extracted biclusters represent combinations of words and documents that form a (latent) topic.  
This is distinguished from traditional LDA topic modeling in that LDA does not cluster documents directly, and words which co-occur across many documents may be clustered even if the shared vocabulary of those documents is small overall.
Instead, a biclustering such as CDP encourages heavy topics which exhibit high co-occurrence of words across documents and documents across words.

The condensed version of the 20 Newsgroup data set is organized into 17 different newsgroups corresponding to four main topics: comp (e.g. computing, graphics), rec (e.g. recreational, sports), sci (e.g. medicine, electronics, space) and talk (e.g. politics, guns), and two smaller topics: religion and miscellaneous for sale. 

CDP found 5 word clusters, 3 news groups, and 3 heavy biclusters. 
There is generally no ground truth for biclusters on text data, and due to the overlapping nature of this "netnews" data set, we chose to evaluate the biclusters by visual inspection. 
We present Table~\ref{tab:text_biclust} showing the words with the highest co-occurrences across documents.
The first grouping is predominantly about space and political topics, while the second grouping is comprised of recreational, religious and medical topics. 
The third heavy bicluster consists of computational topics.
\vskip -0.2in
\begin{table}[H]
\caption{Selection of the top six words with the highest co-occurrence values across the documents.}
\label{tab:text_biclust}
\vskip 0.15in
\begin{center}
\begin{small}
\begin{sc}
\begin{tabular}{lccr}
\toprule
Topic 1 & Topic 2 & Topic 3  \\
\midrule
Mars  & children   & ftp \\
solar & disease    &  fans  \\
technology & bible   &  files  \\
satellite  & baseball &  format  \\
shuttle   & cancer   &  fact      \\
president  & patients   &  games   \\   
\bottomrule
\end{tabular}
\end{sc}
\end{small}
\end{center}
\vskip -0.1in
\end{table}

\subsection{Results for Single cell RNA Sequencing Data}
Biclustering scRNA seq data is commonly used to define developmental stages based solely on the transcriptome in addition to accounting for variation in the data, and identifying biologically important genes and their signatures for each cell stage. 
Each bicluster is an association between groups of cell stages and their genetic drivers. 

A key contribution of CDP is the ability to identify the cell stages and their genetic drivers without having to find highly expressed genes \textit{\'a priori}. 
Furthermore, cell stages are dynamic in time and a probabilistic clustering assignment allows us to capture part of this dynamic without a true time series model. 
This contribution is a vital reason as to why we utilize the $MAP$ to determine the most probable number of clusters instead of running the two DPMMS until they converge on a single value. 

We apply CDP to the scRNA seq data set in \cite{Marques2016OligodendrocyteSystem.}. 
The authors performed a biclustering analysis using BackSPIN \cite{Zeisel2015} and found 13 cell clusters. 

CDP found 7 gene clusters, 12 cell clusters, and 4 strong biclusters.
Like text data, there is generally no ground truth for biclusters on scRNA seq data. 
We evaluate our method using the PANTHER classification system and tools \cite{Mi2018Panther1} \cite{Thomas2006Panther2} \cite{Mi2019Panther3}, and also compare it to \cite{Marques2016OligodendrocyteSystem.}'s results. 

The four biclusters with the strongest co-occurrence values consist of myelin-forming oligodendrocytes (MFOL2), and several stages of mature oligodendrocytes (MOL5, MOL4 and MOL3). 
Biclusters with weaker co-occurrence values consist of newly formed oligodendrocytes (NFOL1) and oligodendrocyte precursor cells (OPC). 
The oligodendrocyte precursor cells can differentiate into newly formed oligodendrocytes, which produce myelin and continue maturing.
Since there are multiple stages of maturation, the composition of the strong biclusters are expected and are corroborated by \cite{Marques2016OligodendrocyteSystem.}. 
The majority of the oligodendrocyte cells are no longer precursor cells or newly formed; they are in differing stages of maturation.

Furthermore, CDP shows that the oligodendrocyte classes also correspond to different regions of the central nervous system. 
For example, oligodendrocytes classified as MFOL2 are also found in abundance in the substantia nigra ventral tegmental (SN-VTA) and hypothalamus regions of the central nervous system. 
Likewise, oligodendrocytes classified as MOL5 are found in abundance in the dorsal horn.

With respect to the genes, CDP did not find distinct gene groupings. 
However, CDP did find two overlapping groupings and multiple groupings with weak co-occurrence values.
Using PANTHER, we find that the two overlapping groupings are strongly affiliated with binding, particularly enzymatic binding, and catalytic activity. 
One group is more involved with cytoskeletal protein binding, and at a higher cellular level, is associated with cellular response to stimulus and cellular metabolic processes. 
The second group is more involved with signaling receptor binding, and with cell component organization and signal transduction at a higher level. 
Genes associated with other biological processes such as the lipid metabolic process or the multicellular organismal process are in the biclusters with weaker co-occurrence values.

\section{Discussion} \label{sec:discussion}
In this paper, we presented a novel, non-parametric probabilistic biclustering method designed to address the challenges of model and parameter selection required by competing methods.  
By utilizing two infinite mixture models and calculating their mutual dependence,  we are able to estimate the number of biclusters strictly from the data and prior, and identify the biclusters without strong modeling assumptions. 

CDP currently requires hyperparameter specifications, but putting a prior on these hyperparameters may improve accuracy without the need for running the model over a range of parameters.
Furthermore, CDP is focused on partitioning discrete data since text and scRNA seq data naturally have count data. 
However, other applications such as audio retrieval do not. 
CDP has the ability to model continuous data as well by changing the multinomial base distribution to a Normal--Inverse--Wishart base distribution and modifying the mutual dependence calculation steps.

Simulation results suggest CDP significantly improves upon DT2B and current standard methods, with more accurate estimation of biclusters, and lower variance estimates.
Experimental results on real data with high sparsity ($>85\%$) demonstrate that CDP is able to extract meaningful heavy biclusters. 
In single cell analyses, this advantage is particularly useful as the data is extremely sparse and noisy.

As a probabilistic model leveraging DPMMs for bicluster estimation, CDP can easily be extended to include additional structure and assumptions.  For instance, in the context of single cell analysis, known results on gene networks may be incorporated through the DPMM priors.  Furthermore, by choosing continuous DPMM base measures $G^r, G^c$, CDP can be applied for biclustering a matrix of continuous values, providing an important advantage over DT2B, which can only accommodate discrete values.

\section{Data and Software}

All data sets are publicly available. 
The condensed 20 Newsgroup data set is available on Sam Roweis's website \cite{RoweisData}. 
The scRNA seq data set is part of the Hemberg lab's collection of publicly available scRNA seq data sets \cite{HembergData} as a SingleCellExperiment Bioconductor S4 class \cite{Lun_SCE_R}. 

We removed rows and columns where the entire vector consisted of zeros. 
For the scRNA seq data set, we also combined the counts of genes that had been split into multiple entries based on loci position. 

Source code for CDP can be found at \href{https://github.com/micnngo/CDP}{https://github.com/micnngo/CDP}.
DPMMs were run using the 'exposed\_parr' branch of DPMMSubClusters \cite{Dinari2019DistributedJulia}. 
The main CDP script is written in R with a wrapper for Julia and C++. 
Plaid, Cheng and Church, Spectral and BiMax algorithms were run using the package 'biclust' in R \cite{biclustR}. The source code for DT2B is available on Github \cite{Rugeles2017Biclustering:Models}.

%\section*{Acknowledgements}

%\textbf{Do not} include acknowledgements in the initial version of
%the paper submitted for blind review.

\section*{Acknowledgements}

This work was supported by NSF grants DMS1936833 and DMS1763272, Simons Foundation grant (594598, QN), and NIH grant R01MH115697. 
We also thank Or Dinari for sharing a modification to his DPMMSubClusters package.


\bibliography{example_paper}
\bibliographystyle{icml2020}


%%%%%%%%%%%%%%%%%%%%%%%%%%%%%%%%%%%%%%%%%%%%%%%%%%%%%%%%%%%%%%%%%%%%%%%%%%%%%%%
%%%%%%%%%%%%%%%%%%%%%%%%%%%%%%%%%%%%%%%%%%%%%%%%%%%%%%%%%%%%%%%%%%%%%%%%%%%%%%%

\end{document}